\documentclass{article}

\PassOptionsToPackage{authoryear,round}{natbib}
\usepackage[preprint]{atomworld_mirror}

\usepackage[utf8]{inputenc}
\usepackage[T1]{fontenc}

\usepackage{amsmath}
\usepackage{amssymb}
\usepackage{amsfonts}
\usepackage{nicefrac}

\usepackage{graphicx}
\usepackage{subcaption}
\usepackage{wrapfig}

\usepackage{booktabs}
\usepackage{makecell}
\usepackage{tabularx}

\usepackage{microtype}
\usepackage{xcolor}
\usepackage{pifont}
\usepackage{placeins}

\usepackage{url}
\usepackage[
    colorlinks=true,
    linkcolor=black,
    citecolor=black,
    urlcolor=blue
]{hyperref}

\usepackage[most]{tcolorbox}

\usepackage{fancyhdr}

\usepackage{colortbl}

\definecolor{AtomWorldRow}{HTML}{EAF2F8}
\definecolor{TableHeader}{HTML}{F4F6F8}
\definecolor{GroupHeader}{HTML}{EEF3F7}

\definecolor{AtomPurple}{HTML}{7B35D5}
\definecolor{AtomBlue}{HTML}{365F9B}
\definecolor{AtomCyan}{HTML}{36A6B8}
\definecolor{AtomRed}{HTML}{B83232}

\definecolor{FrontBoxBlue}{HTML}{F0F4F8}

\makeatletter

\renewcommand{\@toptitlebar}{}
\renewcommand{\@bottomtitlebar}{}

\renewcommand{\@notice}{}

\makeatother

\AtBeginDocument{
  \newgeometry{
    letterpaper,
    left=0.78in,
    right=0.78in,
    top=1.05in,
    bottom=0.72in,
    headheight=0.58in,
    headsep=0.12in,
    footskip=24pt
  }
}

\newtcolorbox{frontmatterbox}{
    enhanced,
    width=\textwidth,
    colback=FrontBoxBlue,
    colframe=FrontBoxBlue,
    boxrule=0pt,
    arc=12pt,
    outer arc=12pt,
    left=10mm,
    right=10mm,
    top=9mm,
    bottom=9mm,
    before skip=0pt,
    after skip=8mm
}

\fancypagestyle{airfirstpage}{
    \fancyhf{}

    \fancyhead[L]{%
        \raisebox{-0.05in}{%
            \includegraphics[
                height=0.48in,
                keepaspectratio
            ]{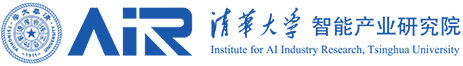}%
        }%
    }

    \fancyfoot{}

}

\usepackage{float}
\usepackage{algorithm}
\usepackage{algpseudocode}
\usepackage{array}
\usepackage{tikz}
\usetikzlibrary{arrows.meta,positioning,fit,backgrounds,calc,tikzmark}
\definecolor{mirrorrowblue}{RGB}{235,242,249}
\setcitestyle{authoryear,round,citesep={;},aysep={,},yysep={;}}

\renewenvironment{abstract}{\par\begingroup\normalfont\fontsize{9.8}{12.1}\selectfont}{\par\endgroup}

\title{AtomWorld-Mirror: Macro-Step World Modeling of Critical Evolution Backbones for Materials Dynamics}

\author{%
  \mbox{\textbf{Ziming Pan}$^{1,7,*}$},
  \mbox{\textbf{Ruge Zhang}$^{2,3,7,*}$},
  \mbox{\textbf{Haozhi Han}$^{4,7,*}$},
  \mbox{\textbf{Junkai Zhou}$^{5}$},
  \mbox{\textbf{Xingyuan Chen}$^{6}$},
  \mbox{\textbf{Yifeng Chen}$^{4}$},
  \mbox{\textbf{Yunquan Zhang}$^{2}$},
  \mbox{\textbf{Ting Cao}$^{7}$},
  \mbox{\textbf{Yunxin Liu}$^{7}$}, and
  \mbox{\textbf{Kun Li}$^{7,\dagger}$}
}
\begin{document}
\thispagestyle{airfirstpage}

\begin{frontmatterbox}
{\raggedright
\fontsize{22.5}{25.5}\selectfont\bfseries
{\color{AtomBlue}Atom}{\color{AtomCyan}World}{\color{AtomRed}-Mirror}: Macro-Step World Modeling of Critical Evolution Backbones for Materials Dynamics
\par}
\vspace{0.46cm}

{\raggedright
\fontsize{10.8}{13.2}\selectfont\bfseries
\makeatletter\@author\makeatother
\par}
\vspace{0.22cm}

{\raggedright\normalfont\fontsize{9.7}{11.9}\selectfont
$^1$Yonsei University, Seoul, Republic of Korea\\
$^2$Institute of Computing Technology, Chinese Academy of Sciences, Beijing, China\\
$^3$University of Chinese Academy of Sciences, Beijing, China\\
$^4$School of Computer Science, Peking University, Beijing, China\\
$^5$Economics \& Technology Research Institute, China National Petroleum Corporation, Beijing, China\\
$^6$Shenzhen Research Institute of Big Data, Shenzhen, China\\
$^7$Institute for AI Industry Research (AIR), Tsinghua University, Beijing, China\\
$^*$Equal contribution.\quad $^\dagger$Corresponding author.\\
\textit{This work was supported by Tecorigin.}
\par}
\vspace{0.42cm}

{\raggedright\normalfont\fontsize{9.2}{11.2}\selectfont
\textbf{Email:} \href{mailto:likun@air.tsinghua.edu.cn}{\textcolor{blue}{likun@air.tsinghua.edu.cn}}\par}
\vspace{0.22cm}

\begin{abstract}

  Atomistic simulation is a fundamental tool for studying long-term materials evolution, from diffusion and defect dynamics to interfacial reactions and fracture. Yet conventional simulators typically advance at microscopic resolution, spending substantial computation on low-impact local updates before reaching structurally consequential states---an {evolutionary-resolution bottleneck} that limits long-horizon simulation. 
  We propose AtomWorld-Mirror, a time-aware macro-step world model for the critical evolution backbone of atomic systems. For Step-Wise atomistic simulation, AtomWorld-Mirror distills short micro-event segments into physically reachable transitions between key states, jointly predicting sparse structural edits and accumulated physical time through latent macro-step dynamics. Local reachability, inventory conservation, and continuous-time consistency constrain each transition. By amortizing local atomic physics into a reusable latent macro model and replacing explicit micro-event replay with macro-step inference, this formulation provides a path toward substantially faster prediction of long-term materials evolution while preserving structural validity and time semantics. Across five atomic systems, spanning Cu-rich RPV steel irradiation aging, Cu--Zr metallic glass, and Li$_3$N-based anti-perovskite solid electrolyte, macro-step inference delivers a speed up of $10^3$ to $10^4$ times over event-by-event simulation.
\end{abstract}
\vspace{0.42cm}
{\normalfont\fontsize{9.2}{11.2}\selectfont
\textbf{Project Page:} \url{https://atomworld-mirror.likun.tech}\par
\textbf{Code:} \url{https://github.com/RSIScience/AtomWorld-Mirror}\par}
\end{frontmatterbox}

\section{Introduction}
\begin{figure}[H]
  \centering
  \includegraphics[width=\linewidth]{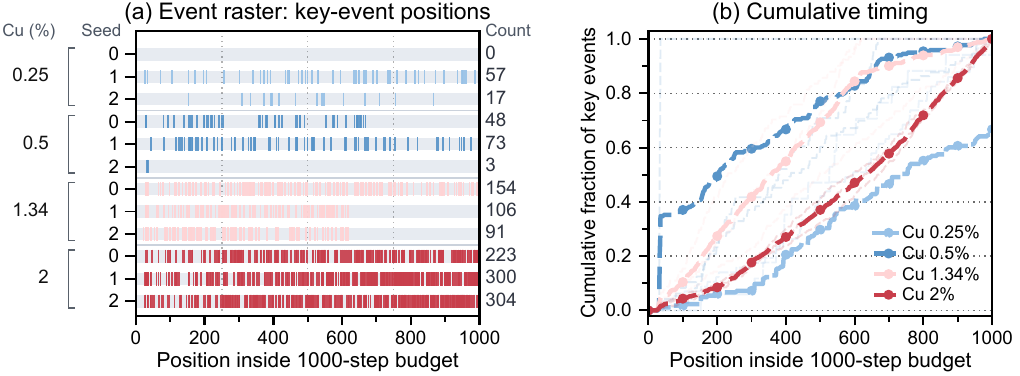}
  \caption{Random timing of key structural evolution within a fixed kinetic simulator replay budget. Each row is one kinetic simulator trajectory at a specified Cu density and random seed; ticks mark Cu-vacancy exchange events within a $1000$-micro-event budget, and the cumulative curves show how these events accumulate across the budget.}
  \label{fig:key-evolution-budget}
\end{figure}

Conventional atomistic simulation methods usually describe materials evolution through explicit time stepping or micro-event updates, e.g., kinetic Monte Carlo (KMC) and molecular dynamics (MD)~\citep{alder1959md,bortz1975nfold}. This Step-Wise formulation treats different state changes uniformly, even though they contribute differently to long-horizon evolution. Some changes are transient fluctuations, whereas others accumulate into persistent structural progress. Figure~\ref{fig:key-evolution-budget} illustrates this mismatch in Cu-rich reactor pressure vessel (RPV) steel aging trajectories, where Cu-vacancy exchange events associated with key structural evolution occur sparsely and at seed-dependent times within the same fixed replay budget. As a result, standard simulators may spend substantial computation on low-impact intermediate states before reaching structurally consequential configurations~\citep{voter2002timescale,soisson1996cufe,soisson2007cufe}. We characterize this mismatch as an evolutionary-resolution bottleneck. Overcoming it requires compressing low-impact intermediate states while retaining the sparse transitions that carry persistent structural progress. We refer to this retained sequence as the critical evolution backbone.

A rollout over this backbone must determine how far the system should advance while preserving both structural progress and the physical time accumulated in between. The underlying atomistic simulator resolves valid local events and their time costs, but it does not directly indicate which later state should be retained next. That state may lie many microscopic events ahead, and both its structural change and duration depend on the intervening path. Rather than resolving this path event by event, long-horizon modeling can compress it into a sequence of macro steps that predicts the next physically reachable backbone transition together with the time required to reach it.

To provide this capability, we propose AtomWorld-Mirror, a time-aware macro-step world model for atomistic evolution. AtomWorld denotes the underlying atomistic simulator, which serves during training as a teacher that provides reference trajectories, local transition rules, and physical-time labels. Mirror distills these microscopic dynamics into a latent macro model that predicts reachable sparse edits and their accumulated physical duration directly from the current configuration. At rollout, it evaluates candidate macro horizons, selects a legal transition, and advances both structure and clock without explicitly replaying the intervening microscopic events~\citep{kingma2014vae,hafner2019planet,hafner2020dreamer,hafner2025dreamer4,fichthorn1991dmc,chatterjee2007kmc}. This yields a compressed but physically constrained state-time rollout for long-horizon structural prediction. To provide this capability, we develop AtomWorld-Mirror, a distillation framework that distills conventional atomistic simulation methods into time-aware macro-step world models.

 {Such compression is useful only if each predicted macro transition remains consistent with the underlying atomistic dynamics, which leads to three physical constraints.} (i) \emph{Local Reachability}. AtomWorld-Mirror restricts structural prediction to candidate supports  {constructed from the local event rules}, ensuring that edits remain within sites accessible to the underlying dynamics. (ii) \emph{Inventory Conservation}. The projection-closed sparse-edit formulation preserves material counts when predicted edits are converted into lattice states. (iii) \emph{Continuous-Time Consistency}. Physical duration is modeled as a path-conditioned quantity because related macro endpoints can accumulate different waiting times along different microscopic paths~\citep{gillespie1976general,gillespie1977exact,fichthorn1991dmc}.  {Accordingly, teacher path summaries supervise duration modeling during training, while prior path latents provide the corresponding information during rollout.}

 {Together, these components allow Mirror to replace microscopic replay with physically constrained macro evolution without discarding structural or temporal information.} Across controlled atomistic simulations, AtomWorld-Mirror captures backbone-level sparse state changes through physically valid edits, aligns path-conditioned time and energy predictions along teacher trajectories, and delivers a speed up of $10^3$ to $10^4$ times in a timed end-to-end diagnostic against event-by-event simulation.

Our contributions are threefold:

\begin{itemize}

\item We introduce a simulator-to-world-model distillation method that converts step-wise atomistic {teacher simulators} into time-aware macro-step models over critical evolution backbones.

\item We propose {AtomWorld-Mirror}, a time-aware macro-step world model distilled from these teacher simulators to predict critical state-time transitions beyond step-wise microscopic replay.

\item Across five atomic systems with KMC and MD teachers, AtomWorld-Mirror delivers physically valid rollouts and $10^3$--$10^4\times$ speedup while preserving structural and temporal accuracy.

\end{itemize}

\section{Related work}

\textbf{Fine-Grained Atomistic Dynamics.}
Atomistic simulators traverse material evolution through different microscopic representations. Molecular dynamics (MD) integrates atomic coordinates through short time steps~\citep{verlet1967fluids,plimpton1995fast}, whereas kinetic Monte Carlo (KMC) samples state-dependent events and advances a residence-time clock~\citep{bortz1975nfold,fichthorn1991dmc}. High-performance implementations such as OpenKMC extend event-driven KMC to very large systems~\citep{li2019openkmc}. In RPV steel aging, neutron irradiation drives the formation of Cu-rich precipitates~\citep{soisson1996cufe,soisson2007cufe}; vacancy-mediated KMC represents this process through explicit local events, transition rates, and physical-time labels.

\textbf{Accelerated Atomistic Dynamics.}
Several lines of work reduce the cost of traversing these trajectories. Superbasin KMC aggregates recurrent low-barrier events within local basins and advances their effective dynamics~\citep{fichthorn2013lskmc}. Rate Scaling KMC reduces the rates of fast events to limit repetitive sampling and accelerate access to slower evolution~\citep{lin2019partialscaling}. Adaptive KMC discovers transition mechanisms and constructs event tables during simulation, extending the accessible dynamics beyond a predefined event catalogue~\citep{henkelman2001akmc,chatterjee2007kmc}. Learned potentials, graph and equivariant simulators, and physics-informed models amortize local force and interaction calculations across atoms~\citep{behler2007nn,bartok2010gap,schutt2017schnet,batatia2022mace,zhang2018deeppotential,raissi2019pinn,pfaff2021meshgraphnets}.

\section{Problem Setup}

We formalize this state-time compression problem starting from the microscopic trajectory of an atomistic simulator.

\subsection{Micro-event resolution and key-state evolution}

RPV steel aging makes the resolution mismatch explicit. Let \(X_t\) denote the atomic configuration after the \(t\)-th microscopic update and \(T_t\) its physical clock. A conventional simulator resolves a fine-grained trajectory \(\{X_t\}_{t=0}^{N}\), where each transition \(X_t\to X_{t+1}\) is a local integration step~\citep{verlet1967fluids,plimpton1995fast,thompson2022lammps} or an atom-vacancy event~\citep{bortz1975nfold,gillespie1977exact}. Each update follows the simulator's local physics, while long-term behavior concentrates in a much smaller set of structurally decisive states. The learning problem is to identify those states and retain the structural progress they carry.

We denote a sparse key-state subsequence by
\begin{equation}
\mathcal{B}=(X_{n_0},X_{n_1},\ldots,X_{n_M}), \quad 0=n_0<n_1<\cdots<n_M\leq N, \quad M\ll N,
\label{eq:backbone-definition}
\end{equation}
 and call it the critical evolution backbone. The retention criterion is system-dependent: RPV steel retains successive Cu-cluster growth and connectivity milestones, and MD cage-swap systems retain topology-change event signatures. The realized macro horizon is $k_m=n_{m+1}-n_m\in\{1,\ldots,1024\}$, so each transition ends at the next important state and adapts to the current configuration.

\subsection{Macro-step state-time objective}

We train this predictive selection over $\mathcal{K}=\{1,\ldots,1024\}$. For each state, the next important state determines the realized macro horizon $k$ and the number of teacher micro events compressed into the transition. The corresponding physical duration $\tau_{t:t+k}$ is predicted jointly with the endpoint. A direct full-state objective is
\begin{equation}
p_\theta(X_{t+k},\tau_{t:t+k}\mid X_t,k),
\end{equation}
where $\tau_{t:t+k}=T_{t+k}-T_t$ for a realized path. Most lattice sites remain unchanged over a macro step, making full-state prediction inefficient. We express the structural target as sparse edit labels over a physically reachable candidate set $\mathcal{C}_k(X_t)$:
\begin{equation}
\Delta X_{t:t+k}=\{(i,a_i)\}_{i\in\mathcal{C}_k(X_t)},
\end{equation}
where $a_i$ is the final edit label or occupancy state at candidate site $i$. Unchanged candidates retain their current occupancy, so the effective edit remains sparse. The horizon $k$ determines how far the teacher dynamics can propagate and consequently changes the candidate set. The resulting objective is
\begin{equation}
p_\theta(\Delta X_{t:t+k},\tau_{t:t+k}\mid X_t,\mathcal{C}_k(X_t),k).
\end{equation}
This objective couples structural support, endpoint edits, and path duration. A macro transition is defined by their joint prediction.

\subsection{KMC supervision of continuous time}

In KMC, the current state $X_t$ has possible events $\mathcal{E}(X_t)=\{e_1,\ldots,e_n\}$ with rates $r_i(X_t)$~\citep{bortz1975nfold,gillespie1976general,fichthorn1991dmc}. The total escape rate is
\begin{equation}
R(X_t)=\sum_i r_i(X_t),
\end{equation}
and event selection and waiting time are
\begin{equation}
p(e_i\mid X_t)=\frac{r_i(X_t)}{R(X_t)}, \qquad \Delta t\sim \mathrm{Exp}(R(X_t)).
\end{equation}
KMC realizes a continuous-time Markov chain (CTMC). The normalized rates determine which local event occurs, and the total escape rate determines how long the system remains in its current state. Each micro event consequently advances both structure and physical time. Compressing several such events into one macro transition requires an endpoint edit and the time accumulated across the intervening path.

The expected waiting time at $X_t$ is $1/R(X_t)$, and the sampled clock update is $T_{t+1}=T_t+\Delta t_t$. For a teacher segment of $k$ events, define
\begin{equation}
\tau_{\mathrm{exp}}(t,k)=\sum_{j=0}^{k-1}\frac{1}{R(X_{t+j})}, \qquad
\tau_{\mathrm{real}}(t,k)=T_{t+k}-T_t=\sum_{j=0}^{k-1}\Delta t_{t+j}.
\end{equation}
Mirror uses $\tau_{\mathrm{exp}}$ as its primary macro-time target, preserving the expected CTMC clock along the teacher path. The realized duration $\tau_{\mathrm{real}}$ records one stochastic draw and serves as an uncertainty diagnostic. Paths that reach similar endpoints through different rate environments can accumulate different durations, making macro time path-conditioned. KMC supplies all parts of this supervision directly: local event semantics, reachable trajectories, and continuous-time labels~\citep{chatterjee2007kmc}.

\section{AtomWorld-Mirror}
\label{sec:method}

AtomWorld-Mirror implements the macro-transition objective as a teacher-student world model for backbone-aware atomistic evolution. AtomWorld denotes the teacher-exposed evolution world, including configurations, local event support, transition rules, and physical clock. Mirror is the student-side macro-step world model over backbone state-time transitions. The teacher supplies short, locally valid state-time segments, and Mirror predicts reachable sparse edits, accumulated duration, and the next macro latent state for selected horizons. The primary instantiation uses atomistic KMC; other Step-Wise atomistic simulators that provide sparse structural states and physical-time supervision can expose the same teacher interface and be distilled into Mirror.

Mirror learns macro state-time transitions between consecutive states of the critical evolution backbone $\mathcal{B}$ of Eq.~\ref{eq:backbone-definition}, distilled from KMC-generated segments. For each candidate macro horizon, it takes the current lattice configuration $X_t$, its reachable candidate set $\mathcal{C}_k(X_t)$, and state-derived context as input. It predicts a sparse lattice edit, the next latent state, the accumulated physical duration, and reward/energy-related quantities. The planner scores the resulting transitions, retains a legal candidate, and projects its edit before execution. The projected state becomes the input to the next macro step. One macro step maps
\begin{equation}
\bigl(X_t,\ \mathcal{C}_k(X_t),\ k\bigr)
\;\longmapsto\;
\bigl(\Delta\hat{X}_{t:t+k},\ \hat{z}_{t+k},\ \hat{\tau}_{t:t+k}\bigr),
\end{equation}
 and rollout applies $\hat{X}_{t+k}=X_t\oplus\Delta\hat{X}_{t:t+k}$ and repeats. Each training example contains an endpoint edit, a microscopic path summary, and two duration targets derived from a teacher segment.

The teacher KMC simulator supplies configurations, local reachable supports, microscopic path summaries, and continuous-time labels. Mirror maps the current lattice state and active candidate patch into latent representations, predicts horizon-conditioned macro dynamics, decodes a sparse edit within the candidate support, and predicts accumulated duration. Projection precedes the latent, reward, and duration losses, giving training and rollout the same projection-closed forward path. Figure~\ref{fig:architecture} shows the pipeline, Table~\ref{tab:implementation-details} in Appendix~\ref{app:implementation} consolidates component-level dimensions, inputs, and outputs, and Appendix~\ref{app:training-details} lists the data splits and filtering rules.

\subsection{Teacher supervision}
\label{sec:teacher-supervision}

Given an initial configuration \(X_t\), the teacher advances the underlying
microscopic dynamics over a short segment of horizon \(k\in\mathcal{K}\),
producing the local trajectory \(\{X_{t+i}\}_{i=0}^{k}\)
~\citep{fichthorn1991dmc,chatterjee2007kmc}.
Each supervised macro sample contains
\begin{equation}
(X_t,k,X_{t+k}, \Delta X_{t:t+k}, \tau_{\mathrm{exp}}, \tau_{\mathrm{real}}, s_{\mathrm{path}}, \mathcal{C}_k).
\end{equation}
Here $\mathcal{K}$ is the controlled Multi-K horizon set, $\Delta X_{t:t+k}$ is the start-to-end sparse edit, $s_{\mathrm{path}}$ summarizes the microscopic path, and $\mathcal{C}_k$ is the reachable candidate set. Rollout receives the current state $X_t$, candidate support $\mathcal{C}_k(X_t)$, horizon $k$, and state-derived context. The endpoint $X_{t+k}$, sparse-edit target, path summary, and duration labels supervise training. The realized duration is
\begin{equation}
\tau_{\mathrm{real}}=\sum_{j=0}^{k-1}\Delta t_{t+j},
\end{equation}
while the primary time target is the path-conditioned accumulated expected time
\begin{equation}
\tau_{\mathrm{exp}}=\sum_{j=0}^{k-1}\frac{1}{R(X_{t+j})}.
\end{equation}
We use $\tau_{\mathrm{exp}}$ as the main duration target because it reflects the expected CTMC clock. $\tau_{\mathrm{real}}$ remains an auxiliary stochastic target and time-uncertainty diagnostic.

The three physical commitments stated in Section~1 turn these supervised segments into Mirror rollouts: local reachability restricts edits to the teacher-derived candidate support, inventory conservation constrains the projection that converts latent predictions into lattice edits, and continuous-time consistency makes duration path-conditioned~\citep{gillespie1976general,gillespie1977exact,fichthorn1991dmc}. These commitments enter Mirror through latent dynamics, sparse-edit projection, and duration losses.

\subsection{Mirror macro-step world model}

Mirror is a macro-step world model developed from the DreamerV4 latent dynamics architecture~\citep{hafner2025dreamer4} and adapted to teacher-supervised atomistic state-time transitions. Its encoder combines a latent-variable representation with graph message passing and equivariant site features~\citep{kingma2014vae,gilmer2017mpnn,satorras2021egnn},
\begin{equation}
z_t=\mathrm{Enc}_\theta(X_t,g_t),
\end{equation}
where $g_t$ may contain local composition, defect statistics, rate summaries, vacancy information, or other global context. The active-patch encoder also produces site embeddings for $\mathcal{C}_k(X_t)$, coupling global latent dynamics with the local edit space. The same $X_t$ can lead to different reachable endpoints, and each $k$ defines a distinct admissible path family. AtomWorld-Mirror represents this variation with a horizon-conditioned path latent $h_t$.

During training, the posterior path latent conditions on the endpoint and teacher path summary,
\begin{equation}
q_\phi(h_t\mid z_t,z_{t+k},s_{\mathrm{path}},k),
\end{equation}
which identifies the supervised transition. During inference, the model samples or selects from a prior
\begin{equation}
p_\theta(h_t\mid z_t,g_t,k).
\end{equation}
The macro dynamics then predicts a horizon-conditioned next latent,
\begin{equation}
\hat{z}_{t+k}=z_t+F_\theta(z_t,h_t,k).
\end{equation}
A KL term transfers posterior information into the rollout prior. At rollout, every input is computed from the current state, state-derived context, and macro horizon.

\subsection{Reachability-constrained sparse edits}
\label{sec:reachability-edits}

The edit decoder operates on $\mathcal{C}_k(X_t)$, the reachable candidate support generated by the teacher dynamics. Candidate-restricted event supports are standard in spatial and adaptive KMC~\citep{chatterjee2007kmc,henkelman2001akmc}, with related barrier and path-sampling context in NEB and time-reversal path-sampling work~\citep{henkelman2000neb,liu2023trps}. For candidate site $i$, the decoder takes site embedding $u_i$, patch context, predicted macro latent $\hat{z}_{t+k}$, path latent $h_t$, and horizon $k$, and predicts change logit $\ell_i$ and final type distribution $\pi_i$,
\begin{equation}
(\ell_i,\pi_i)=D_\theta(u_i,c_{k,i},\hat{z}_{t+k},h_t,k),
\end{equation}
Let $m_i=\mathbf{1}\{x_{t+k,i}\neq x_{t,i}\}$ be the teacher change mask and $a_i$ be the target final type. A compact sparse-edit loss is
\begin{equation}
\mathcal{L}_{\mathrm{edit}}
=
-\sum_{i\in\mathcal{C}_k(X_t)}
\left[
m_i\log\sigma(\ell_i)
+(1-m_i)\log(1-\sigma(\ell_i))
+\beta m_i\log \pi_i(a_i)
\right],
\end{equation}
where $c_{k,i}$ is the active-patch context, the first two terms learn sparse support, the last learns changed-site type, and $\beta=1$ in the reported experiments. This candidate-restricted output space focuses prediction on sites that the local dynamics can affect and supplies the support for physically reachable vacancy-mediated edits~\citep{soisson1996cufe,soisson2007cufe}.

Inventory conservation further requires the candidate patch to preserve material counts:
\begin{equation}
\sum_{i\in\mathcal{C}_k(X_t)}\mathbf{1}\{\hat{x}_{t+k,i}=c\}
=
\sum_{i\in\mathcal{C}_k(X_t)}\mathbf{1}\{x_{t,i}=c\},
\qquad c\in\{\mathrm{Fe},\mathrm{Cu},\mathrm{Vac}\},
\end{equation}
Outside the patch, sites are copied from the current configuration. Projection is implemented as paired vacancy--atom transport, selecting vacancy-to-atom and atom-to-vacancy pairs under the budget
\begin{equation}
d_{\mathrm{vac}}\leq k,\qquad d_{\mathrm{Cu}}\leq k,\qquad d_{\mathrm{vac}}+d_{\mathrm{Cu}}\leq 2k,\qquad |\Delta\hat{X}_{t:t+k}|\leq 2k,
\end{equation}
where distances use the periodic BCC hop metric. Projection is applied before latent, reward, and duration losses are evaluated. The projected state is re-encoded and passed to the next macro step, aligning optimization with the state consumed during rollout.

\subsection{Continuous-time duration modeling and loss}
\label{sec:duration}

Duration is a core macro-transition output because stochastic simulation couples event selection and waiting time through the same transition rates~\citep{gillespie1976general,gillespie1977exact,fichthorn1991dmc,lin2019partialscaling}. We separate path-conditioned expected CTMC duration from realized sampled duration: $\tau_{\mathrm{exp}}$ is the main macro-time target, while $\tau_{\mathrm{real}}$ is auxiliary stochastic supervision.

\noindent The duration branch predicts expected macro time in log space,
\begin{equation}
b_k(X_t)=\log k-\log R(X_t), \qquad
\mu_{\mathrm{exp}}=b_k(X_t)+r^\tau_\theta(z_t,\hat{z}_{t+k},h_t,g_t,\rho_t,k),
\end{equation}
where $b_k(X_t)$ is the log start-state CTMC estimate $k/R(X_t)$, $\rho_t$ is edit/projection context, and $r^\tau_\theta$ is a residual. The residual corrects the start-state estimate using rate changes accumulated along the teacher path. The expected-time head captures both the initial escape-rate scale and path-dependent dynamics. It also predicts $\log\sigma_{\mathrm{exp}}$, and the main duration loss is the implementation's log-space Gaussian negative log-likelihood:
\begin{equation}
\mathcal{L}_\tau
=
\log\sigma_{\mathrm{exp}}
+\frac{1}{2}
\left(
\frac{\log(\tau_{\mathrm{exp}}+\epsilon)-\mu_{\mathrm{exp}}}
{\sigma_{\mathrm{exp}}}
\right)^2 .
\end{equation}
Realized duration remains stochastic even under identical rates, so it is modeled as a conditional lognormal auxiliary target:
\begin{equation}
\mu_{\mathrm{real}}=b_k(X_t)+r^{\mathrm{real}}_\theta(z_t,\hat{z}_{t+k},h_t,g_t,\rho_t,k),\qquad
\tau_{\mathrm{real}}\sim \mathrm{LogNormal}(\mu_{\mathrm{real}},\sigma_{\mathrm{real}}).
\end{equation}
Default training uses prior-side duration losses, matching the quantities available during rollout. We use smooth-$L_1$ regression for $\mathcal{L}_{\mathrm{reward}}$ and $\mathcal{L}_{\mathrm{latent}}$. The projection consistency loss $\mathcal{L}_{\mathrm{proj}}$ aligns the re-encoded projected state with the teacher endpoint latent, and $\mathcal{L}_{\mathrm{real}}$ is the lognormal negative log-likelihood of $\tau_{\mathrm{real}}$. The implementation objective is
\begin{equation}
\begin{aligned}
\mathcal{L}
=&\;
\lambda_{\mathrm{edit}}\mathcal{L}_{\mathrm{edit}}
+\lambda_{\mathrm{reward}}\mathcal{L}_{\mathrm{reward}}
+\lambda_{\mathrm{latent}}\mathcal{L}_{\mathrm{latent}}
+\lambda_{\mathrm{proj}}\mathcal{L}_{\mathrm{proj}}\\
&+\lambda_{\mathrm{KL}}D_{\mathrm{KL}}(q_\phi\parallel p_\theta)
+\lambda_{\tau}\mathcal{L}_{\tau}
+\lambda_{\mathrm{real}}\mathcal{L}_{\mathrm{real}} .
\end{aligned}
\end{equation}
The macro horizon is selected independently at each step from $\mathcal{K}=\{1,\ldots,1024\}$. The model encodes $X_t$, draws $h_t$ from the prior, and decodes a projected sparse edit for each candidate $k$. Candidates violating reachability or inventory are discarded, and the surviving legal candidates are ranked by predicted structural progress per unit physical time:
\begin{equation}
k^\star=\arg\max_{k\in\mathcal{K}_{\mathrm{legal}}}\;
\frac{\hat{r}_\theta(z_t,\hat{z}_{t+k},k)}{\hat{\tau}_{t:t+k}+\epsilon},
\label{eq:horizon-selection}
\end{equation}
The selected edit is applied as $\hat{X}_{t+k}=X_t\oplus\Delta\hat{X}_{t:t+k}$, the duration $\hat{\tau}_{t:t+k}$ advances the clock, and the step repeats to form a backbone trajectory.

The backbone $\mathcal{B}$ of Eq.~\ref{eq:backbone-definition} remains stable across realized macro horizons and trajectory seeds. We measure sensitivity over $k\in\{1,2,4,\ldots,1024\}$ micro steps on 16 held-out trajectories and compute the persistence ratio, defined as the fraction of backbone events surviving across horizons. Persistence stays above $0.98$ throughout. Across trajectory seeds, the largest-Cu-cluster identity reaches $0.9066$, showing that the persistent structural hierarchy remains stable under microscopic stochasticity.

\section{Experimental design}

\subsection{Experimental scope}

Across five atomic systems, Mirror maintains zero reachability violations and zero inventory violations, and delivers an end-to-end speed up of $10^3$ to $10^4$ times. A reachability violation is an edit outside the candidate set $\mathcal{C}_k(X_t)$, and an inventory violation is a projected patch that changes any species count; both are reported as the fraction of macro steps affected. The evaluation begins with binary RPV steel aging and extends to 3-element and 10-element RPV steels under the same KMC teacher. Cu--Zr metallic glass and Li$_3$N-based anti-perovskite solid electrolyte use MD as the teacher, with cage-swap topology events defining their macro-step boundaries. The primary experiments instantiate this interface with KMC, and Appendix~\ref{app:generality} reports all per-system results, including the MD-teacher systems that reuse the same Mirror architecture and training loop with tuple construction adapted to topology-change events.

The RPV steel systems center the study on alloy aging, where Cu-rich precipitate formation under irradiation controls the degradation of nuclear-reactor pressure vessels~\citep{soisson1996cufe,soisson2007cufe,li2019openkmc}. Cu--Zr metallic glass extends the evaluation to an amorphous alloy with cage-swap dynamics and structural relaxation~\citep{cheng2011atomic}. The Li$_3$N-based anti-perovskite solid electrolyte extends it to ionic migration in a solid-electrolyte composition for lithium batteries~\citep{cordier1989licasrn,dembitskiy2025litraj,zhao2012lisnsr}. Together, these systems test state-time macro modeling across composition, structural order, and simulator type, while the measured speed up supports longer aging, diffusion, and parameter-sweep studies.

AtomWorld-Mirror uses a state-time macro transition as its prediction unit. It identifies structurally decisive waypoints along the critical evolution backbone and distills segments from a traditional atomistic simulator teacher into sparse reachable edits, next latent states, and accumulated physical duration. Each transition retains the local support and inventory conditions required for a valid rollout, while macro-step inference bypasses explicit replay of intervening micro-events. We apply this teacher interface to event-driven KMC, MD-generated topology-change segments in Cu--Zr metallic glass~\citep{cheng2011atomic}, and Li$_3$N-based anti-perovskite solid electrolyte~\citep{cordier1989licasrn,dembitskiy2025litraj,zhao2012lisnsr}, as well as to other Step-Wise atomistic simulators.

\begin{figure}[H]
  \centering
  \includegraphics[width=\linewidth]{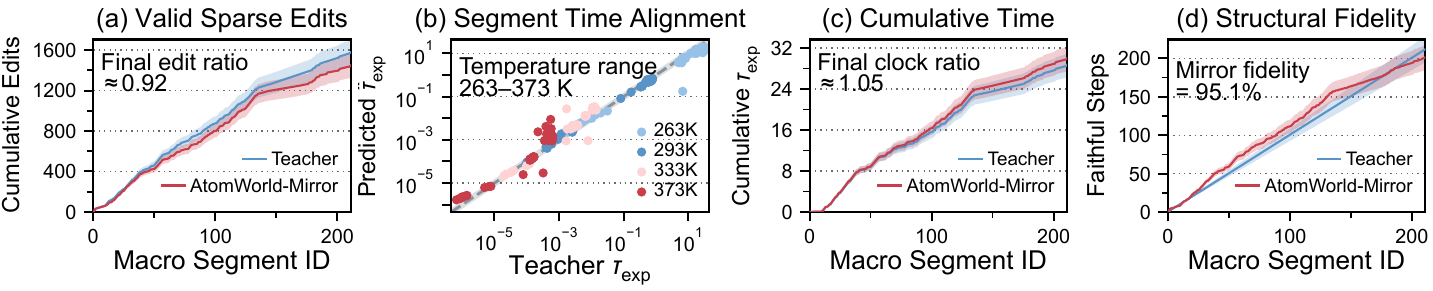}
  \caption{Controlled macro-step validation against the teacher simulator. The panels show (a) cumulative structural edits, (b) single-segment expected-time alignment across temperatures, (c) long-trajectory cumulative expected time, and (d) cumulative correctly typed edits for Teacher and Mirror.}
  \label{fig:main-results}
\end{figure}

\subsection{Event-driven instantiation}

\begin{figure}[H]
  \centering
  \includegraphics[width=\linewidth]{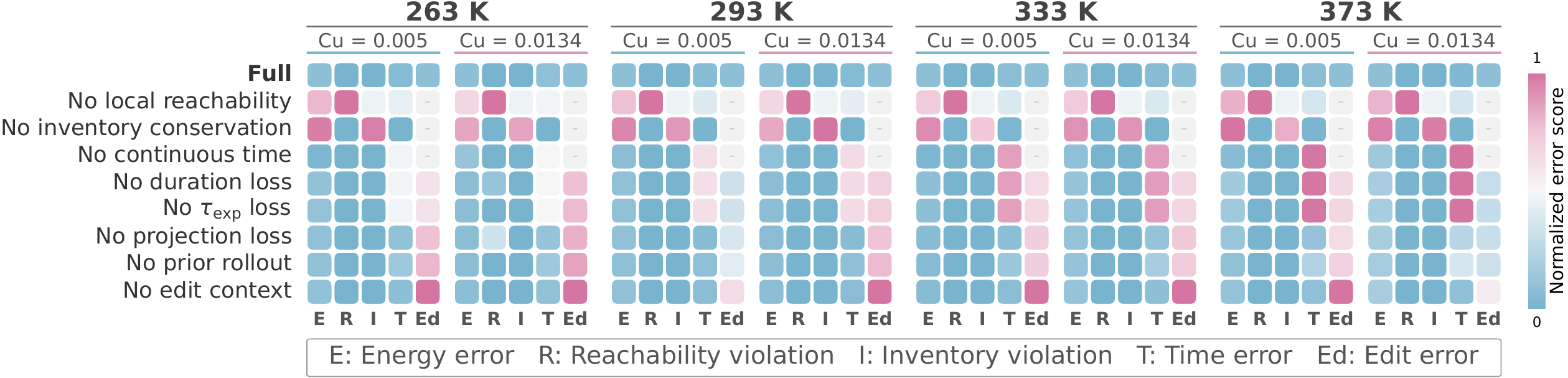}
  \caption{Cu-density and temperature ablation matrices for Multi-K teacher-probe rollout. Darker cells denote larger values of the corresponding failure metric.}
  \label{fig:closed-loop-results}
\end{figure}

We instantiate AtomWorld as an atom-vacancy KMC process to evaluate projected edit validity, expected-duration alignment, closed-loop rollout consistency, and timed macro-inference efficiency.

Unless otherwise stated, the controlled physical setup uses a $40^3$ BCC lattice, Cu density $0.0134$, vacancy density $0.0002$, a 2NN event neighborhood, $16$ defect-graph shells, and reward scale $10.0$. The quantitative matrix uses adaptive $k\in\{1,\ldots,1024\}$ and horizon-selected macro diagnostics. Figure~\ref{fig:main-results} summarizes the controlled macro-step validation against the KMC teacher/reference.

Given these paired macro segments, the first evaluation separates raw prediction quality from projection-closed physical validity. Structural quality is summarized by the edit F1 score, the harmonic mean of precision and recall between predicted and teacher change masks over $\mathcal{C}_k(X_t)$, and by type accuracy, the fraction of teacher-changed sites whose predicted species matches $a_i$. The expected-time head is evaluated against the path-conditioned CTMC target $\tau_{\mathrm{exp}}$ using log-MAE, log-RMSE, log-correlation, and scale ratio:
\begin{equation}
\mathrm{logMAE}_\tau
=
\frac{1}{N}\sum_{n=1}^{N}
\left|\log \hat{\tau}^{(n)}_{\mathrm{exp}}-\log \tau^{(n)}_{\mathrm{exp}}\right|,
\qquad
\mathrm{scale}_\tau
=
\frac{1}{N}\sum_{n=1}^{N}
\frac{\hat{\tau}^{(n)}_{\mathrm{exp}}}{\tau^{(n)}_{\mathrm{exp}}}.
\label{eq:segment-time-metrics}
\end{equation}
Scale ratio tests systematic time bias, log-correlation tests relative variation across teacher segments, and log-RMSE weights the largest per-segment errors. The stochastic $\tau_{\mathrm{real}}$ head is evaluated separately as an auxiliary lognormal using NLL, probability-integral-transform diagnostics, and 68\%/95\% interval coverage, the fraction of segments whose realized duration falls inside the predicted interval. Appendix~\ref{app:training-details} reports training parameters, compute resources, and runtime settings.

\subsection{Closed-loop validity and ablations}

The model maintains an autonomous state $\hat{X}_m$ by applying the previous projected sparse edit. The evaluated model trajectory is the closed-loop sequence
\(\hat{X}_0 \xrightarrow{k_0} \hat{X}_1 \xrightarrow{k_1} \cdots
\xrightarrow{k_{M-1}} \hat{X}_M\),
while each transition is scored against a KMC probe started from the current model state. This protocol evaluates whether the self-generated macro trajectory remains locally reachable, inventory-preserving, temporally calibrated, and physically plausible under local probes.

Figure~\ref{fig:closed-loop-results} reports Multi-K teacher-probe ablation matrices across Cu concentrations $0.005$ and $0.0134$ and temperatures $263$, $293$, $333$, and $373$K. Shared ablation rows measure candidate-support validity, inventory preservation, path-conditioned duration, and the components that maintain editable states during rollout. These matrices use two non-learned references. The KMC teacher supplies the rate-governed state-time target, and the copy-state baseline keeps the lattice fixed while using the CTMC start-state duration from $R(X_t)$. Removing local reachability measures candidate-support dependence, removing inventory conservation measures material-count preservation, and replacing learned duration with a start-state CTMC baseline measures the contribution of path-conditioned information.

\begin{figure}[H]
  \centering
  \includegraphics[width=\linewidth]{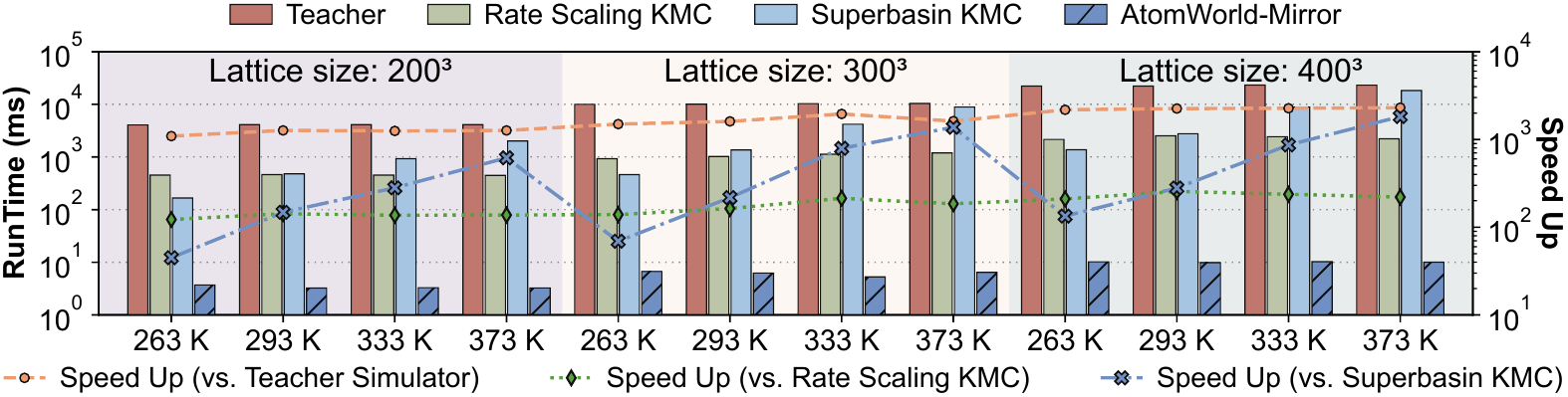}
  \caption{End-to-end timing diagnostic across lattice sizes and temperatures.}
  \label{fig:speedup-benchmark}
\end{figure}

\subsection{Long-horizon diagnostics and efficiency}
Replacing explicit micro-event replay with macro-step inference delivers a speed up of $10^3$ to $10^4$ times across lattice size, temperature, and material system (Fig.~\ref{fig:speedup-benchmark}, Appendix~\ref{app:generality}, Appendix~\ref{app:extreme-scale}). The timing includes candidate construction, projection, benchmark orchestration, and batched neural inference for one adaptive macro transition. The diagnostic measures replay reduction under reachable, inventory-preserving, and temporally calibrated edits, with the reference cost set by classical and KMC simulators~\citep{plimpton1995fast,thompson2022lammps,li2019openkmc}.

Macro-step cost tracks the candidate patch defined by the event neighborhood. Inference cost stays near-constant in system size, and a model trained on $40^3$ runs on larger configurations without retraining. Across Multi-K steps with $k$ ranging from $1$ to $1024$, one macro step replaces $480$ teacher micro events on average and advances the clock by $12.98$ seconds of expected time; Appendix~\ref{app:scale} reports this timing.

Capturing Cu-rich precipitate evolution in RPV steel requires following vacancy-mediated kinetics across large spatial domains and long physical timescales. We evaluate a $50$-year autonomous closed-loop rollout of Mirror on the full $5.4\times10^{10}$-atom RPV steel aging system across four temperatures in Appendix~\ref{app:extreme-scale}.

The teacher-forced diagnostic measures state-conditioned time and reward alignment along a contiguous KMC trajectory $\{X_{t_m}\}_{m=0}^{M}$, where $X_{t_m}\xrightarrow[\mathrm{teacher}]{k_m}X_{t_{m+1}}$. The readout accumulates predicted and reference expected times and the final clock-scale ratio:
\begin{equation}
\small
\hat{T}^{\mathrm{exp}}_M=\sum_{m=0}^{M-1}\hat{\tau}_{\mathrm{exp},m},
\qquad
T^{\mathrm{exp}}_M=\sum_{m=0}^{M-1}\tau_{\mathrm{exp},m},
\qquad
\rho_T=\frac{\hat{T}^{\mathrm{exp}}_M}{T^{\mathrm{exp}}_M+\epsilon}.
\label{eq:teacher-forced-time-ratio}
\end{equation}

\section{Conclusion}

We introduced a simulator-to-world-model distillation method that converts
step-wise atomistic teachers into time-aware macro-step models over critical
evolution backbones. AtomWorld-Mirror instantiates this framework by predicting
reachable sparse edits and path-conditioned physical time while preserving
material inventory. Across five atomic systems with KMC and MD teachers, it
achieves physically valid rollouts and $10^3$--$10^4\times$ speedups over
event-by-event simulation.

\bibliography{references}
\bibliographystyle{atomworld_mirror}

\newpage
\appendix

\section{Training details and parameters}
\label{app:training-details}

\begin{table}[H]
  \centering
  \setlength{\tabcolsep}{3pt}
  \renewcommand{\arraystretch}{1.12}
  \begin{tabular}{@{}>{\raggedright\arraybackslash}m{0.17\linewidth} >{\raggedright\arraybackslash}m{0.28\linewidth} >{\raggedright\arraybackslash}m{0.18\linewidth} >{\raggedright\arraybackslash}m{0.27\linewidth}@{}}
    \toprule
    \textbf{Data / Teacher} & \textbf{Value} & \textbf{Student / Optim.} & \textbf{Value} \\
    \midrule
    \textbf{Teacher} & atom-vacancy KMC & \textbf{Student} & Mirror macro-step world model \\
    \textbf{Lattice} & $40^3$ BCC & \textbf{Candidates} & $\mathcal{C}_k(X_t)$; 8 V seeds \\
    \textbf{Cu / V density} & 0.0134 / 0.0002 & \textbf{Path input} & teacher path summary \\
    \textbf{Temperature} & 300 K & \textbf{Latents} & global $32{\times}16$; patch/path 64/32 \\
    \textbf{Event support} & 2NN hops & \textbf{Epochs / batch} & 120 / 32 \\
    \textbf{Train / val} & 2000 / 400 segments per $k$ & \textbf{Optimizer} & Adam, lr $10^{-4}$ \\
    \textbf{Split policy} & independent envs & \textbf{Weight decay} & $10^{-5}$ \\
    \textbf{Filtering} & no-op skipped & \textbf{Expected / realized time} & 1.0 / 0.25 \\
    \textbf{Time targets} & $\tau_{\mathrm{exp}}$ main; $\tau_{\mathrm{real}}$ aux & \textbf{Edit / reward / latent / proj.} & 1.0 / 0.5 / 0.5 / 0.5 \\
    \textbf{Aux. losses} & proj. / reward / priors & \textbf{KL / prior edit / prior latent} & 0.05 / 0.25 / 0.25 \\
    \textbf{Compute} & NVIDIA A100 GPUs & \textbf{Runtime} & 35.5 min train; 23.3 / 47.1 s eval \\
    \bottomrule
  \end{tabular}
  \caption{Training and evaluation configuration.}
  \label{tab:training-config}
\end{table}

The reported training and evaluation runs use NVIDIA A100 GPUs. Timing values are taken from persisted run logs for the reported configuration.

\section{Implementation details by component}
\label{app:implementation}

\begin{figure}[H]
  \centering
  \scriptsize
  \resizebox{\linewidth}{!}{%
  \begin{tikzpicture}[
    node distance=3.6mm,
    box/.style={draw, rounded corners=1.4pt, align=center, inner sep=2.6pt, minimum height=7.2mm, font=\scriptsize},
    tbox/.style={box, fill=blue!7},
    sbox/.style={box, fill=orange!9},
    obox/.style={box, fill=green!8},
    ar/.style={-{Latex[length=1.5mm]}, thick, gray!70!black},
    lab/.style={font=\tiny\itshape, text=black!65}
  ]
  \node[tbox] (kmc) {AtomWorld teacher (KMC)\\ \tiny rates $r_i$, escape rate $R(X_t)$, clock $T_t$};
  \node[tbox, right=6mm of kmc] (seg) {Macro segment\\ \tiny $(X_t,k,X_{t+k},\Delta X,\tau_{\mathrm{exp}},\tau_{\mathrm{real}},s_{\mathrm{path}},\mathcal{C}_k)$};
  \draw[ar] (kmc) -- (seg);

  \node[sbox, below=9mm of kmc.south west, anchor=north west, text width=25mm] (enc) {Graph encoder\\ \tiny $z_t=\mathrm{Enc}_\theta(X_t,g_t)$\\ \tiny patch embeddings $u_i$};
  \node[sbox, right=4mm of enc, text width=24mm] (path) {Path latent $h_t$\\ \tiny prior $p_\theta(h_t\mid z_t,g_t,k)$\\ \tiny posterior $q_\phi(\cdot\mid z_{t+k},s_{\mathrm{path}})$};
  \node[sbox, right=4mm of path, text width=23mm] (dyn) {Macro dynamics\\ \tiny $\hat{z}_{t+k}=z_t+F_\theta(z_t,h_t,k)$};
  \draw[ar] (enc) -- (path);
  \draw[ar] (path) -- (dyn);
  \draw[ar] (seg.south) -- ++(0,-2mm) -| (path.north);

  \node[obox, below=8mm of enc.south west, anchor=north west, text width=25mm] (edit) {Sparse-edit decoder\\ \tiny $(\ell_i,\pi_i)$ on $\mathcal{C}_k$};
  \node[obox, right=4mm of edit, text width=24mm] (dur) {Duration head\\ \tiny $\mu_{\mathrm{exp}}=b_k+r^\tau_\theta$};
  \node[obox, right=4mm of dur, text width=23mm] (rew) {Reward / energy head\\ \tiny $\hat{r}_\theta$};
  \draw[ar] (dyn.south) -- ++(0,-2.5mm) -| (edit.north);
  \draw[ar] (dyn) -- (dur);
  \draw[ar] (dyn.south east) -- ++(0,-2.5mm) -| (rew.north);

  \node[box, fill=red!7, below=8mm of dur, text width=52mm] (proj) {Inventory projection (paired vacancy--atom transport)\\ \tiny budget $d_{\mathrm{vac}}\!+\!d_{\mathrm{Cu}}\leq 2k$, $|\Delta\hat{X}|\leq 2k$};
  \draw[ar] (edit.south) -- ++(0,-2mm) -| (proj.north west);
  \draw[ar] (dur) -- (proj);
  \node[box, fill=red!7, below=5mm of proj, text width=52mm] (sel) {Horizon selection over $k\in\mathcal{K}_{\mathrm{legal}}$\\ \tiny $k^\star=\arg\max_k \hat{r}_\theta/(\hat{\tau}+\epsilon)$};
  \draw[ar] (proj) -- (sel);
  \node[box, fill=gray!8, right=5mm of sel, text width=27mm] (next) {Next macro state\\ \tiny $\hat{X}_{t+k}=X_t\oplus\Delta\hat{X}$, $T\!+\!=\!\hat{\tau}$};
  \draw[ar] (sel) -- (next);
  \draw[ar] (next.east) -- ++(5mm,0)
    |- ([xshift=1.5mm,yshift=3mm]enc.north east)
    -| (enc.north);
  \node[lab, anchor=west] at ($(next.east)+(6mm,3mm)$) {closed loop};

  \node[lab, left=1mm of enc, rotate=90, anchor=south] {student};
  \node[lab, left=1mm of kmc, rotate=90, anchor=south] {teacher};
  \end{tikzpicture}%
  }
  \caption{AtomWorld-Mirror architecture. The teacher supplies segments with reachable candidate supports, path summaries, and continuous-time labels. Mirror encodes the configuration and active patch, draws a horizon-conditioned path latent from the posterior during training and the prior during rollout, predicts the macro latent, and decodes a sparse edit, accumulated duration, and reward/energy quantity. Inventory projection precedes the downstream losses, so the projected state consumed during rollout is also the state used for optimization.}
  \label{fig:architecture}
\end{figure}
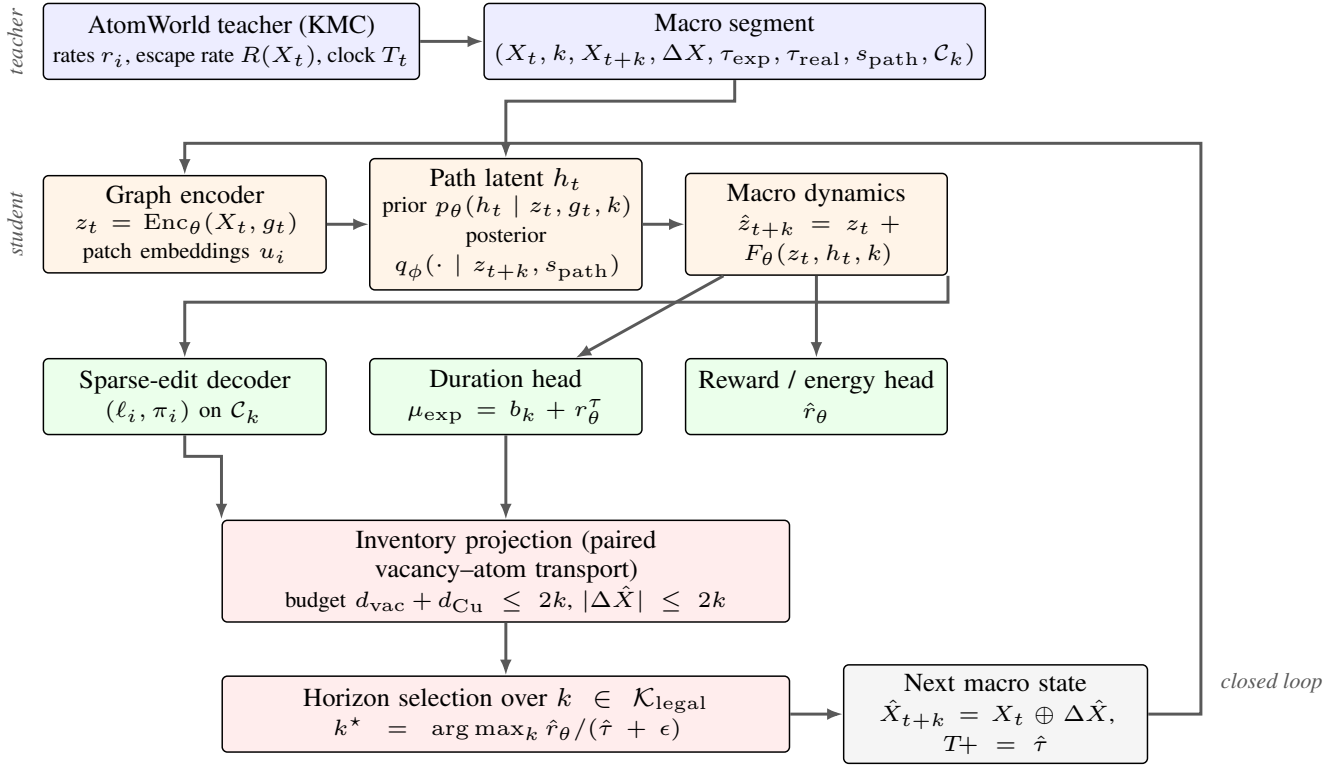

Figure~\ref{fig:architecture} shows the full pipeline, and Table~\ref{tab:implementation-details} consolidates the component-level inputs, outputs, and dimensions used throughout Section~4.

\begin{table}[H]
  \centering
  \small
  \setlength{\tabcolsep}{4pt}
  \renewcommand{\arraystretch}{1.2}
  \begin{tabular}{@{}>{\raggedright\arraybackslash}p{0.19\linewidth} >{\raggedright\arraybackslash}p{0.26\linewidth} >{\raggedright\arraybackslash}p{0.26\linewidth} >{\raggedright\arraybackslash}p{0.19\linewidth}@{}}
    \toprule
    \textbf{Component} & \textbf{Input} & \textbf{Output} & \textbf{Rollout-time} \\
    \midrule
    Graph encoder & $X_t$, global context $g_t$ & global latent $z_t$ ($32{\times}16$), site embeddings $u_i$ & yes \\
    Candidate builder & $X_t$, 2NN event support, $k$ & $\mathcal{C}_k(X_t)$ from 8 vacancy seeds & yes \\
    Path posterior $q_\phi$ & $z_t$, $z_{t+k}$, $s_{\mathrm{path}}$, $k$ & path latent $h_t$ (dim 32) & no (training only) \\
    Path prior $p_\theta$ & $z_t$, $g_t$, $k$ & path latent $h_t$ (dim 32) & yes \\
    Macro dynamics $F_\theta$ & $z_t$, $h_t$, $k$ & residual to $\hat{z}_{t+k}$ & yes \\
    Edit decoder $D_\theta$ & $u_i$, patch context (dim 64), $\hat{z}_{t+k}$, $h_t$, $k$ & change logit $\ell_i$, type distribution $\pi_i$ over \{Fe, Cu, Vac\} & yes \\
    Inventory projection & raw edit, $\mathcal{C}_k$, budget $2k$ & projection-closed $\Delta\hat{X}_{t:t+k}$ & yes \\
    Duration head & $z_t$, $\hat{z}_{t+k}$, $h_t$, $g_t$, edit context $\rho_t$, $k$ & $\mu_{\mathrm{exp}}$, $\log\sigma_{\mathrm{exp}}$ on top of $b_k(X_t)$ & yes \\
    Realized-time head & same as duration head & $\mu_{\mathrm{real}}$, $\sigma_{\mathrm{real}}$ (lognormal) & yes \\
    Reward head & $z_t$, $\hat{z}_{t+k}$, $k$ & $\hat{r}_\theta$ (reward scale $10.0$) & yes \\
    Horizon selector & $\{\hat{r}_\theta,\hat{\tau}\}_{k\in\mathcal{K}_{\mathrm{legal}}}$ & $k^\star$ by Eq.~\ref{eq:horizon-selection} & yes \\
    \bottomrule
  \end{tabular}
  \caption{Component-level implementation details. Rollout-time availability marks whether a component's inputs are computable from $X_t$ and $k$ alone.}
  \label{tab:implementation-details}
\end{table}

\section{Macro-step pseudocode}
\label{app:pseudocode}

Algorithm~\ref{alg:mirror-rollout} restates the rollout forward path of Section~\ref{sec:method} as executable pseudocode. Colored comments mark the three stages the implementation separates: latent encoding of the configuration and active patches, per-horizon prediction followed by inventory projection, and horizon selection with clock advance. Every quantity consumed inside the loop is computable from the current state and the horizon alone, the contract recorded in the rollout-time column of Table~\ref{tab:implementation-details}.

\begin{algorithm}[H]
\caption{Mirror macro-step rollout over the critical evolution backbone. Colored comments mark encoding, projection-closed prediction, and horizon selection with clock advance.}
\label{alg:mirror-rollout}
\begin{algorithmic}[1]
\Require configuration $X_0$, clock $T_0$, trained Mirror $\theta$, horizon set $\mathcal{K}=\{1,\ldots,1024\}$, candidate builder $\mathcal{C}_k(\cdot)$
\Ensure backbone trajectory $\mathcal{B}$
\State $t \gets 0$; $m \gets 0$; $X_t \gets X_0$; $T \gets T_0$; $\mathcal{B} \gets [(X_0,T_0)]$
\While{the rollout budget remains}
  \State \textcolor{teal!60!black}{\textit{// encode}}\quad $z_t \gets \mathrm{Enc}_\theta(X_t, g_t)$; site embeddings $u_i$ per site
  \State $\mathcal{K}_{\mathrm{legal}} \gets \emptyset$
  \State \textbf{for} $k \in \mathcal{K}$ \textbf{do}
  \State \hspace{1.5em}$\mathcal{C}_k \gets \mathrm{CandidateSupport}(X_t,k)$; construct active-patch context $c_{k,i}$
  \State \hspace{1.5em}$h_k \sim p_\theta(h \mid z_t, g_t, k)$;\quad $\hat{z}_{t+k} \gets z_t + F_\theta(z_t, h_k, k)$
  \State \hspace{1.5em}$(\ell_i, \pi_i) \gets D_\theta(u_i,c_{k,i},\hat{z}_{t+k},h_k,k)$ over $i \in \mathcal{C}_k$
  \State \hspace{1.5em}$\Delta\hat{X}_k \gets \mathrm{Decode}(\ell, \pi)$;\quad \textcolor{red!60!black}{\textit{// inventory projection}}
  \State \hspace{1.5em}$\Delta\hat{X}_k \gets \mathrm{Project}(\Delta\hat{X}_k,\mathcal{C}_k,2k)$ \hfill \textcolor{blue!60!black}{$\triangleright$ paired vacancy--atom transport}
  \State \hspace{1.5em}$\hat{r}_k \gets \mathrm{Reward}_\theta(z_t,\hat{z}_{t+k},k)$; $\hat{\tau}_k \gets \mathrm{Duration}_\theta(z_t,\hat{z}_{t+k},h_k,g_t,k)$
  \State \hspace{1.5em}\textbf{if} $\Delta\hat{X}_k$ satisfies reachability and inventory \textbf{then} add $k$ to $\mathcal{K}_{\mathrm{legal}}$ \textbf{else} discard $k$ \textbf{end if}
  \State \textbf{end for}
  \State \textbf{if} $\mathcal{K}_{\mathrm{legal}}=\emptyset$ \textbf{then} record the stop reason and \textbf{break}
  \State \textcolor{violet!60!black}{\textit{// horizon selection}}\quad $k^\star \gets \arg\max_{k \in \mathcal{K}_{\mathrm{legal}}} \hat{r}_k/(\hat{\tau}_k+\epsilon)$
  \State $X_t \gets X_t \oplus \Delta\hat{X}_{k^\star}$; $T \gets T+\hat{\tau}_{k^\star}$; $t \gets t+k^\star$; $m \gets m+1$; append $(X_t,T)$ to $\mathcal{B}$
\EndWhile
\State \Return $\mathcal{B}$
\end{algorithmic}
\end{algorithm}

\section{Generality across material systems and teacher types}
\label{app:generality}

The four additional systems extend the three physical commitments across chemistry, bonding character, and teacher type. RPV steel keeps the atom-vacancy KMC teacher of Section~\ref{sec:teacher-supervision} and raises the element count to 3 (Fe--Cu--Ni) and 10 (adding Mn, Cr, Al, Si, P, Co, Mo). Cu--Zr metallic glass and the Li$_3$N-based anti-perovskite solid electrolyte use molecular dynamics, with a macro step ending at a cage-swap topology event and its duration given by the elapsed MD time between consecutive events. The MD systems carry no vacancy species. A snap-to-lattice bijection places one atom at each reference site, conserving inventory by construction; species-level edits appear as paired swaps, the direct analog of KMC Cu--vacancy transport.

Each table reports the teacher as a reference alongside AtomWorld-Mirror. The decoder restricts edits to $\mathcal{C}_k$, and the projection conserves species counts. The time columns report log-MAE and the per-segment scale ratio of Eq.~\ref{eq:segment-time-metrics}.

\subsection{RPV steel, multi-component (KMC teacher)}

\begin{table}[H]
  \centering
  \small
  \setlength{\tabcolsep}{6pt}
  \renewcommand{\arraystretch}{1.15}
  \begin{tabular}{llcc}
    \toprule
    \textbf{System} & \textbf{Model} & \textbf{violation\% $\cdot$ F1} & \textbf{log-MAE $\cdot$ scale} \\
    \midrule
    RPV 3-element  & KMC    & 0\% $\cdot$ 1.00 & 0.00 $\cdot$ 1.00 \\
    \rowcolor{mirrorrowblue}             & \textbf{Mirror} & \textbf{0\% $\cdot$ 0.77} & \textbf{0.21 $\cdot$ 1.05} \\
    \addlinespace[2pt]
    RPV 10-element & KMC    & 0\% $\cdot$ 1.00 & 0.00 $\cdot$ 1.00 \\
    \rowcolor{mirrorrowblue}             & \textbf{Mirror} & \textbf{0\% $\cdot$ 0.78} & \textbf{0.32 $\cdot$ 1.06} \\
    \bottomrule
  \end{tabular}
  \caption{RPV steel: physical validity and time calibration at 3 and 10 elements.}
  \label{tab:rpv-physicality}
\end{table}

\begin{figure}[H]
  \centering
  \includegraphics[width=\linewidth]{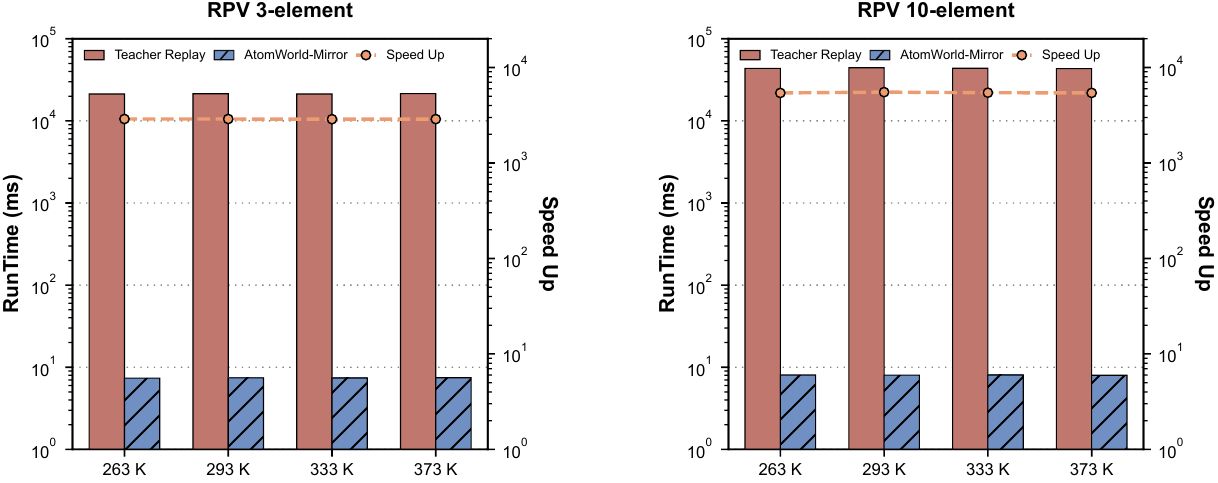}
  \caption{End-to-end speed up over the KMC teacher for the RPV steel systems at 263, 293, 333, and 373 K. Bars show runtime and the line shows speed up.}
  \label{fig:rpv-speedup}
\end{figure}

Increasing the RPV composition from 3 to 10 elements roughly doubles the speed up. Added chemistry raises the teacher's per-event cost, while macro-step cost remains tied to the candidate patch.

\subsection{Cu--Zr metallic glass (MD teacher)}

\begin{table}[H]
  \centering
  \small
  \setlength{\tabcolsep}{6pt}
  \renewcommand{\arraystretch}{1.15}
  \begin{tabular}{@{}lll@{}}
    \toprule
    \textbf{Model} & \textbf{violation\% $\cdot$ F1} & \textbf{log-MAE $\cdot$ scale} \\
    \midrule
    MD teacher & 0\% $\cdot$ 1.00 & 0.00 $\cdot$ 1.00 \\
    \multicolumn{1}{>{\columncolor{mirrorrowblue}[\tabcolsep][\tabcolsep]}l}{\hspace*{-\tabcolsep}\textbf{Mirror}} &
    \multicolumn{1}{>{\columncolor{mirrorrowblue}[\tabcolsep][\tabcolsep]}l}{\textbf{0\% $\cdot$ 0.84}} &
    \multicolumn{1}{>{\columncolor{mirrorrowblue}[\tabcolsep][\tabcolsep]}l}{\textbf{0.26 $\cdot$ 1.10}} \\
    \bottomrule
  \end{tabular}
  \caption{Cu--Zr metallic glass: physical validity and time calibration.}
  \label{tab:cuzr-physicality}
\end{table}

\begin{figure}[H]
  \centering
  \includegraphics[width=\linewidth]{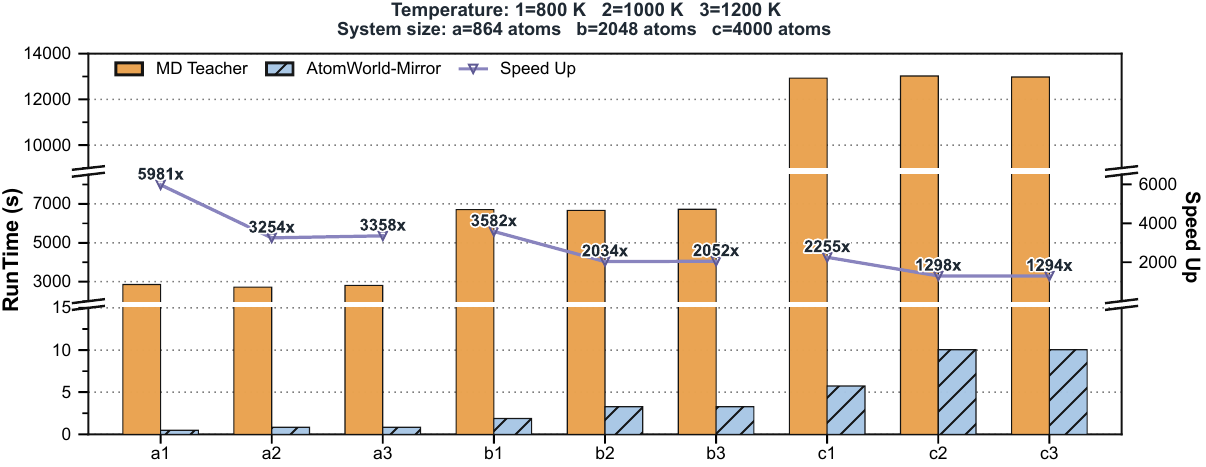}
  \caption{End-to-end speed up over the MD teacher for the Cu--Zr metallic glass across temperature and system size.}
  \label{fig:cuzr-speedup}
\end{figure}

Cu--Zr provides the largest speed up in this study, with edit F1 $0.84$ and duration scale $1.10$. The MD teacher resolves femtosecond integration steps between rare cage swaps, allowing each macro step to replace a long stretch of integration. Higher temperature increases cage-swap frequency and shortens the interval covered by one macro step.

\subsection{Li$_3$N-based anti-perovskite solid electrolyte (MD teacher)}

\begin{table}[H]
  \centering
  \small
  \setlength{\tabcolsep}{6pt}
  \renewcommand{\arraystretch}{1.15}
  \begin{tabular}{@{}lll@{}}
    \toprule
    \textbf{Model} & \textbf{violation\% $\cdot$ F1} & \textbf{log-MAE $\cdot$ scale} \\
    \midrule
    MD teacher & 0\% $\cdot$ 1.00 & 0.00 $\cdot$ 1.00 \\
    \multicolumn{1}{>{\columncolor{mirrorrowblue}[\tabcolsep][\tabcolsep]}l}{\hspace*{-\tabcolsep}\textbf{Mirror}} &
    \multicolumn{1}{>{\columncolor{mirrorrowblue}[\tabcolsep][\tabcolsep]}l}{\textbf{0\% $\cdot$ 0.71}} &
    \multicolumn{1}{>{\columncolor{mirrorrowblue}[\tabcolsep][\tabcolsep]}l}{\textbf{0.27 $\cdot$ 1.07}} \\
    \bottomrule
  \end{tabular}
  \caption{Li$_3$N-based anti-perovskite solid electrolyte: physical validity and time calibration.}
  \label{tab:linsr-physicality}
\end{table}

\begin{figure}[H]
  \centering
  \includegraphics[width=\linewidth]{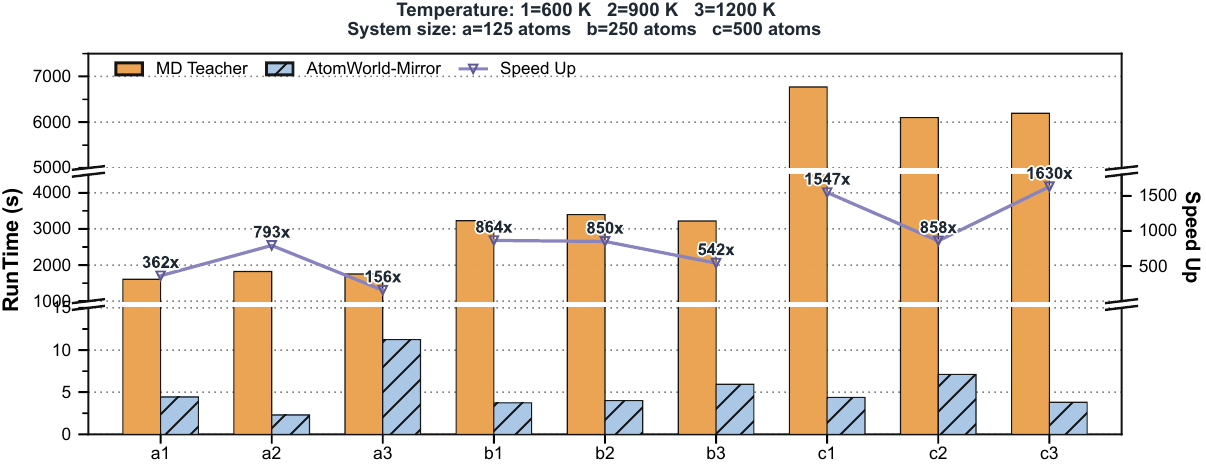}
  \caption{End-to-end speed up over the MD teacher for the Li$_3$N-based anti-perovskite solid electrolyte across temperature and system size.}
  \label{fig:linsr-speedup}
\end{figure}

Mirror reaches edit F1 $0.71$ on the electrolyte, with duration scale $1.07$. Ionic migration produces well-separated topology events, keeping the duration target sharp.

\section{Separating the macro-step formulation from learned dynamics}
\label{app:constraint-baselines}

We isolate the macro-transition formulation and its three physical constraints through three matched comparisons.

The first comparison holds the latent backbone fixed while removing the macro-step structure. A Dreamer-style variant omits reachability, inventory, and continuous-time constraints. A direct endpoint regressor predicts $X_{t+k}$ over the full lattice without candidate restriction, path latent, or duration head, reusing the main model's graph encoder for representation parity.

\begin{table}[H]
  \centering
  \small
  \setlength{\tabcolsep}{5pt}
  \renewcommand{\arraystretch}{1.15}
  \begin{tabular}{lccc}
    \toprule
    \textbf{Model} & \textbf{Physically valid sparse edits} & \textbf{log-MAE $\cdot$ scale} & \textbf{$\rho_T$} \\
    \midrule
    Teacher KMC                    & violation 0\% $\cdot$ F1 1.00 & 0.00 $\cdot$ 1.00 & 1.00 \\
    Dreamer-style (no constraints) & reach 45.2\% $\cdot$ inv 59.2\% & 4.08 $\cdot$ 0.35 & 0.35 \\
    Endpoint regression            & change-F1 0.42 $\cdot$ type-acc 0.45 & N/A & N/A \\
    \rowcolor{mirrorrowblue} \textbf{AtomWorld-Mirror} & \textbf{violation 0\% $\cdot$ F1 0.91} & \textbf{0.13 $\cdot$ 1.00} & \textbf{1.01} \\
    \bottomrule
  \end{tabular}
  \caption{Constraint and formulation baselines against AtomWorld-Mirror on the shared RPV steel aging held-out split (Cu $0.0134$, $300$ K, $k\in\{1,\ldots,1024\}$). The table quantifies the contributions of reachability, inventory conservation, and duration modeling.}
  \label{tab:constraint-baselines}
\end{table}

At matched computational cost, the constrained macro-step formulation delivers reachable edits, inventory preservation, and calibrated physical time. AtomWorld-Mirror matches the Dreamer-style variant within $0.5\%$ in wall-clock time, raises reachable edits from $45.2\%$ to $100\%$, and raises the cumulative clock ratio from $0.35$ to $1.01$. AtomWorld-Mirror reaches change-F1 $0.91$ with path-conditioned duration modeling, compared with $0.42$ for the endpoint regressor, which has no duration output.

\section{Teacher-forced long-trajectory protocol}
\label{app:teacher-forced}

At segment $m$, from teacher state $X_{t_m}$, the model predicts a projected macro candidate. When several horizons are legal, the planner selects one using Eq.~\ref{eq:horizon-selection}. The teacher advances the matching microscopic dynamics to provide the reference next state, reward, expected time, and realized time. The model contributes $(\Delta\hat{X}_m,\hat{\tau}_{\mathrm{exp},m},\hat{r}_m)$, and the evaluator reports cumulative expected-time curves, selected-horizon histograms, stop reasons, and cumulative energy-error summaries. The teacher-forced diagnostic uses the teacher state as input at every segment, isolating duration-scale error, horizon-selection bias, and reward misalignment from autonomous rollout drift. The per-segment scale ratio of Eq.~\ref{eq:segment-time-metrics} is the time-calibration summary reported in the per-system tables of Appendix~\ref{app:generality}. The cumulative clock ratio $\rho_T$ of Eq.~\ref{eq:teacher-forced-time-ratio} is reported in Table~\ref{tab:constraint-baselines} and Table~\ref{tab:extreme-scale-validity}.

\section{Adaptive horizon selection}
\label{app:adaptive-horizon}

At each macro step, the selector ranks legal horizons after projection. The resulting histogram records local configuration difficulty through selected horizons and projection stop events. When every candidate is rejected, the segment is recorded with its stop reason.

\section{Macro-step inference over $k\in\{1,\ldots,1024\}$}
\label{app:scale}

Macro-step inference cost is set by the candidate patch defined by the event neighborhood. Cost per macro step stays near-constant in the total number of atoms, and a checkpoint trained at $40^3$ applies to larger configurations without retraining.

We measure the prior forward pass plus inventory projection on held-out active patches drawn from KMC trajectories over the operating range $k\in\{1,\ldots,1024\}$. One macro step covers $480$ teacher micro events on average and advances the physical clock by $12.98$ seconds of accumulated expected time. Table~\ref{tab:longhorizon-inference} reports representative measurements within this range.

\begin{table}[H]
  \centering
  \small
  \setlength{\tabcolsep}{6pt}
  \renewcommand{\arraystretch}{1.1}
  \begin{tabular}{@{}lrrrr@{}}
    \toprule
    \textbf{Horizon $k$} & \textbf{128} & \textbf{256} & \textbf{512} & \textbf{1024} \\
    \midrule
    Coverage               & 0.889 & 0.906 & 0.780 & 0.649 \\
    \bottomrule
  \end{tabular}
  \\[4pt]
  \begin{tabular}{@{}lr@{}}
    \toprule
    \textbf{Aggregate measurement} & \textbf{Value} \\
    \midrule
    Samples / batch size                       & 96 / 32 \\
    Wall time per pass                         & 134.7 s \\
    Throughput                                 & 0.713 segments/s \\
    Mean micro events per macro step           & 480.0 \\
    Mean expected duration per macro step      & 12.98 s \\
    \bottomrule
  \end{tabular}
  \caption{Measured macro-step inference within the operating range $k\in\{1,\ldots,1024\}$ on NVIDIA A100 GPUs. The table reports representative horizons; coverage is the fraction of attempted segments that yield a usable macro sample after terminal and no-op filtering.}
  \label{tab:longhorizon-inference}
\end{table}

Longer horizons replace more micro-event replay per macro step. Coverage changes from $0.889$ at $k=128$ to $0.649$ at $k=1024$, making the efficiency-coverage relationship explicit.

\section{54-billion-atom RPV steel aging benchmark}
\label{app:extreme-scale}

We evaluate a $50$-year autonomous closed-loop rollout of Mirror on the full $5.4\times10^{10}$-atom RPV steel aging system at $1.34\,\mathrm{at.}\%$ Cu concentration across four temperatures. OpenKMC runs on $5.2$ million cores of the Sunway TaihuLight supercomputer, whose many-core processors are based on SW26010, while Mirror runs on NVIDIA A100 GPUs. The benchmark records physical validity, cumulative time alignment, and end-to-end speed up against OpenKMC.

\begin{table}[H]
  \centering
  \small
  \setlength{\tabcolsep}{5pt}
  \renewcommand{\arraystretch}{1.15}
  \begin{tabular}{lccc}
    \toprule
    \textbf{Model} & \textbf{Physically valid edits} & \textbf{Single-segment time} & \textbf{Cumulative clock} \\
    \midrule
    OpenKMC & violation 0\% $\cdot$ F1 1.00 & log-MAE 0.00 $\cdot$ scale 1.00 & $\rho_T$ 1.00 \\
    \rowcolor{mirrorrowblue} \textbf{Mirror} & \textbf{violation 0\% $\cdot$ F1 0.91} & \textbf{log-MAE 0.27 $\cdot$ scale 1.17} & \textbf{$\rho_T$ 0.89} \\
    \bottomrule
  \end{tabular}
  \caption{Physical validity and time alignment at the $5.4\times10^{10}$-atom scale in the RPV steel aging benchmark.}
  \label{tab:extreme-scale-validity}
\end{table}

At the $5.4\times10^{10}$-atom scale, Mirror retains physically valid edits with F1 $0.91$ and reaches a cumulative clock ratio $\rho_T=0.89$.

\begin{figure}[!htbp]
  \centering
  \includegraphics[width=\linewidth]{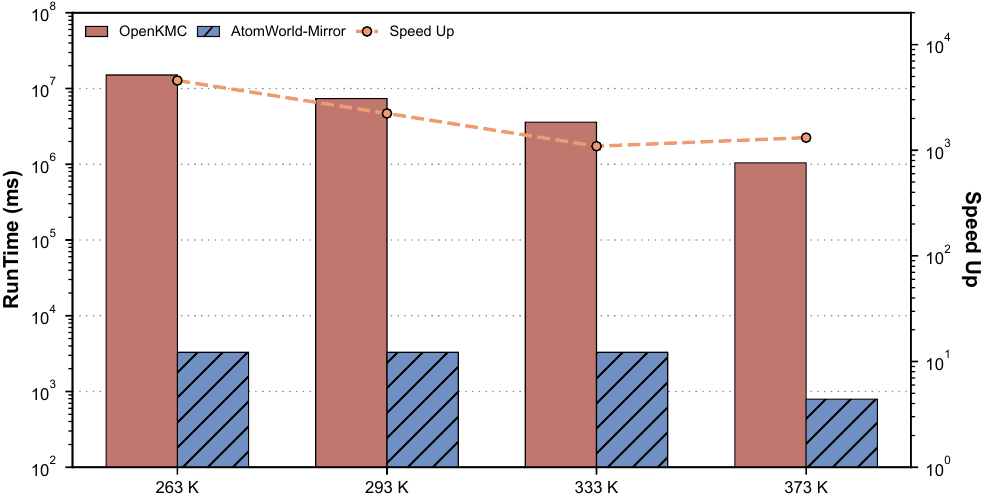}
  \caption{End-to-end speed up of Mirror over OpenKMC for the $5.4\times10^{10}$-atom RPV steel aging benchmark over a $50$-year physical timescale. Bars show runtime and the line shows speed up.}
  \label{fig:extreme-scale-speedup}
\end{figure}

\section{RPV steel aging visualization}
\label{app:cu-cluster-evolution}

\begin{figure}[!htbp]
  \centering
  \includegraphics[width=0.92\linewidth]{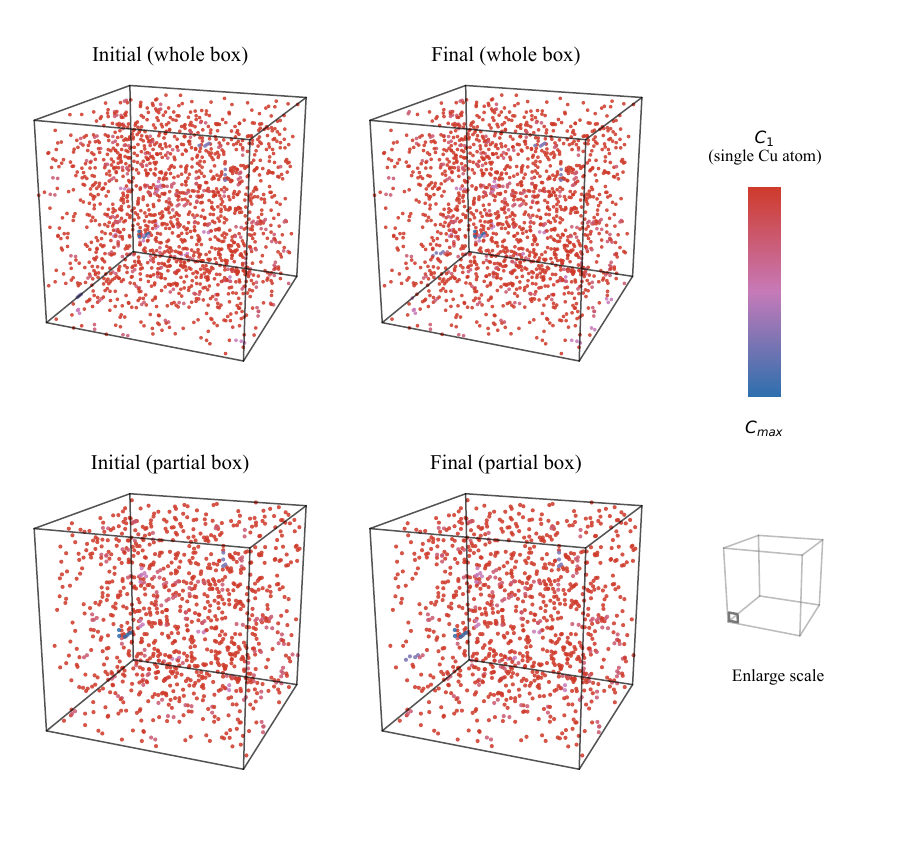}
  \caption{Cu-cluster visual evolution from initial to final states in an RPV steel aging simulation. Each point denotes a Cu atom in the teacher-simulated lattice. Colors use a shared red--pink-purple--blue scale for the actual 1NN/2NN-connected Cu cluster size $C_i$, with red denoting single-Cu clusters and blue denoting the maximum cluster size in this trajectory.}
  \label{fig:cu-cluster-evolution}
\end{figure}

Figure~\ref{fig:cu-cluster-evolution} shows Cu-cluster aging in a vacancy-mediated KMC trajectory on a $40^3$ BCC lattice. The full boxes show the simulated lattice, and the partial boxes enlarge a representative sub-volume to resolve local cluster growth. Colors encode Cu-cluster sizes computed on the full periodic BCC lattice from the teacher configuration.

\section{Limitations and future work}
\label{app:limitations}

AtomWorld-Mirror targets long-horizon materials evolution in regimes where explicit microscopic replay consumes most of the computational budget before structurally consequential states are reached. The study evaluates this setting with controlled simulator teachers and establishes extensions to additional teacher types and materials systems. Future work will extend the framework to broader simulator-teacher combinations and optimize distributed deployment.

The five-system evaluation spans binary and multi-component RPV steel aging, Cu--Zr metallic glass, and Li$_3$N-based anti-perovskite solid electrolyte across KMC and MD teachers. The controlled RPV steel aging setting establishes the primary edit benchmark, and the MD systems demonstrate transfer of the state-time interface across bonding character and teacher type. The current operating range reaches $k=1024$. Extending candidate-support construction and conservation projection to systems with larger species inventories, together with topology criteria for continuously evolving structures, defines the next methodological step.

\section{Expanded related-work discussion}
\label{app:expanded-related-work}

Materials simulation spans several representations of physical evolution. Coarse-grained molecular dynamics, coarse-grained KMC, Monte Carlo, and Markov state models reduce atomic, event, or state resolution through coarse variables, aggregated events, or metastable-state representations~\citep{marrink2007martini,noid2008mscg,garnier2013cgkmc,chodera2014msm}. Classical molecular dynamics advances atomic configurations through short integration steps~\citep{alder1959md,rahman1964argon,verlet1967fluids,plimpton1995fast,thompson2022lammps}. High-throughput materials databases~\citep{jain2013materialsproject,curtarolo2012aflow} and structure/property analysis toolchains~\citep{ong2013pymatgen,ward2018matminer,choudhary2020jarvis} standardize structures, computed properties, and analysis workflows. These abstractions reduce computational cost. AtomWorld-Mirror adds a predictive backbone representation that selects future structural waypoints and assigns the accumulated physical duration of each sparse transition.

Learning-based materials simulation develops reusable models of energies, forces, local structures, and interacting-system dynamics~\citep{merchant2023gnome}. Neural-network and Gaussian approximation potentials~\citep{behler2007nn,bartok2010gap}, graph neural networks for molecules and materials~\citep{schutt2017schnet,xie2018cgcnn,batatia2022mace,gilmer2017mpnn,gasteiger2020dimenet}, and equivariant force fields~\citep{batzner2022nequip,musaelian2023allegro} amortize local physical computations across atoms and molecules. Deep potential molecular dynamics and related packages turn these learned local models into scalable dynamical simulators~\citep{zhang2018deeppotential,wang2018deepmdkit}. Physics-informed learning~\citep{raissi2019pinn,karniadakis2021piml} and graph-network simulators~\citep{battaglia2016interaction,kipf2018nri,pfaff2021meshgraphnets} extend learned dynamics toward differential-equation, mesh-based, and interacting-system settings. AtomWorld-Mirror complements these approaches with a sparse macro state-time prediction target that explicitly represents backbone transitions and accumulated physical duration.

Step-Wise atomistic simulation provides the continuous-time teacher used in our experiments. KMC samples local events from physical rates and advances a residence-time clock~\citep{bortz1975nfold,gillespie1977exact,fichthorn1991dmc}, supplying local event semantics, reachable trajectories, and continuous-time labels~\citep{chatterjee2007kmc}. High-performance KMC systems such as OpenKMC use engineered parallel execution to reach very large systems~\citep{li2019openkmc}. Established KMC accelerators include superbasin KMC and adaptive or self-learning KMC, which consolidate repeated events or construct transition mechanisms during simulation~\citep{fichthorn2013lskmc,henkelman2001akmc}. Parallel-replica, temperature-accelerated, rate-scaling, and policy-guided methods further alter event traversal or event selection~\citep{voter1998parallelreplica,sorensen2000tad,zamora2016spectad,lin2019partialscaling,schulman2017ppo,tang2024tks,bojesen2018pgmc}. These methods operate on event-level traversal, event aggregation, or trajectory selection. AtomWorld-Mirror operates at the macro-transition level: the simulator teacher supplies the state-time reference, and Mirror learns sparse backbone transitions with accumulated physical duration. speed up follows from this predictive representation of atomistic dynamics.

Rate Scaling KMC reduces the sampling frequency of fast event channels by scaling their rates, allowing a fixed event budget to cover more of the slower evolution~\citep{lin2019partialscaling}. Event selection and waiting times are then computed from the scaled rates, so the resulting clock follows the modified process; agreement with the original kinetics depends on the scaling regime and the observable being evaluated. This approach retains explicit event-level updates and targets the cost of repeatedly sampling fast dynamics. AtomWorld-Mirror changes the prediction unit to a reachable macro edit with a jointly predicted duration, amortizing the intervening event sequence into a learned state-time transition.

Table~\ref{tab:accelerator-taxonomy} organizes the comparison by the quantity changed and the physical clock preserved. Trajectory accelerators modify the accepted event sequence. Analytical accelerators retain event-level execution and aggregate fast events inside basins. Rate-scaling methods retain event-level execution while modifying fast-channel rates and their associated waiting-time distribution. AtomWorld-Mirror predicts a sparse edit between backbone states, assigns its accumulated duration as a path-conditioned quantity, and enforces reachability and inventory at every macro step. Structural and temporal quantities remain explicit at the macro level.

\begin{table}[H]
  \centering
  \scriptsize
  \setlength{\tabcolsep}{4pt}
  \renewcommand{\arraystretch}{1.0}
  \begin{tabular}{@{}>{\raggedright\arraybackslash}p{0.2\linewidth} >{\raggedright\arraybackslash}p{0.24\linewidth} >{\raggedright\arraybackslash}p{0.24\linewidth} >{\raggedright\arraybackslash}p{0.22\linewidth}@{}}
    \toprule
    \textbf{Family} & \textbf{What it changes} & \textbf{Physical clock} & \textbf{Examples} \\
    \midrule
    Trajectory accelerators & the accepted event sequence & not directly traceable once event order is altered & RL/PPO policies for KMC, neural surrogate event samplers~\citep{schulman2017ppo,tang2024tks,bojesen2018pgmc} \\
    Analytical accelerators & rates or grouping of fast events; event-level execution retained & preserved, every effective event still resolved & superbasin KMC, adaptive KMC~\citep{fichthorn2013lskmc,henkelman2001akmc} \\
    Rate-scaling accelerators & fast-event rates; event-level execution retained & sampled from scaled rates; fidelity to original kinetics depends on scaling regime & Rate Scaling KMC~\citep{lin2019partialscaling} \\
    \textbf{AtomWorld-Mirror} & \textbf{the prediction unit: sparse macro edit between backbone states} & \textbf{predicted explicitly as path-conditioned accumulated duration} & \textbf{this work} \\
    \bottomrule
  \end{tabular}
  \caption{Where AtomWorld-Mirror sits among KMC acceleration families.}
  \label{tab:accelerator-taxonomy}
\end{table}

\end{document}